\documentclass[10pt,twocolumn,letterpaper]{article}

\usepackage{iccv}
\usepackage{times}
\usepackage{epsfig}
\usepackage{graphicx}
\usepackage{amsmath}
\usepackage{amssymb}

\def\eg{{\emph{e.g.}}}
\def\ie{{\emph{i.e.}}}

\def\etal{{\emph{et al.}}}
\usepackage{graphicx}
\usepackage{amsmath}
\usepackage{amssymb}
\usepackage{booktabs}
\usepackage{pifont} 
\usepackage{amsmath}
\usepackage{amssymb}
\usepackage{graphicx}
\usepackage{array}
\usepackage{comment}
\usepackage{epsfig}
\usepackage{times}
\usepackage{comment}
\usepackage{url}
\usepackage{subfig}
\usepackage{times}
\usepackage{algorithm,algorithmic,xspace}
\usepackage{verbatim}
\usepackage{multirow}
\usepackage{colortbl}
\usepackage{bbding}
\usepackage{pifont}
\usepackage{wasysym}
\usepackage{amssymb}
\usepackage[table]{xcolor}
\usepackage{breqn}
\usepackage{graphicx,animate}
\usepackage[normalem]{ulem}
\usepackage{tikz}
\newcommand*{\circled}[1]{\lower.7ex\hbox{\tikz\draw (0pt, 0pt)%
    circle (.4em) node {\makebox[1em][c]{\small #1}};}}

\newcommand{\Tref}[1]{Table~\ref{#1}}
\newcommand{\Eref}[1]{Equation~(\ref{#1})}
\newcommand{\Fref}[1]{Figure~\ref{#1}}
\newcommand{\Sref}[1]{Section~\ref{#1}}

\def\etal{\emph{ et al.}}
\def\ie{\emph{i.e.}}
\def\eg{\emph{e.g.}}

\def\eg{{\emph{e.g.}}}
\def\ie{{\emph{i.e.}}}

\def\etal{{\emph{et al.}}}

\def\M{{\mathbf{M}}}

\def\Pi{{\mathbf{M}_{i}}}

\def\CopyNeRF{CopyRNeRF~}

\definecolor{Auqamarin}{gray}{0.9}
\definecolor{LightCyan}{rgb}{0.88,1,1}
\definecolor{myGreen}{rgb}{0, .9, .6}

\definecolor{americanrose}{rgb}{1.0, 0.01, 0.24}

\makeatletter
\newcommand*{\rom}[1]{\expandafter\@slowromancap\romannumeral #1@}
\DeclareRobustCommand\onedot{\futurelet\@let@token\@onedot}
\def\@onedot{\ifx\@let@token.\else.\null\fi\xspace}

\makeatletter
  \newcommand\figcaption{\def\@captype{figure}\caption}
  \newcommand\tabcaption{\def\@captype{table}\caption}
\makeatother

\usepackage[breaklinks=true,bookmarks=false]{hyperref}

\usepackage[capitalize]{cleveref}
\crefname{section}{Sec.}{Secs.}
\Crefname{section}{Section}{Sections}
\Crefname{table}{Table}{Tables}
\crefname{table}{Tab.}{Tabs.}
\iccvfinalcopy 

\def\iccvPaperID{6271} 

\ificcvfinal\fi

\begin{document}

\title{CopyRNeRF: Protecting the CopyRight of Neural Radiance Fields}

\author{Ziyuan Luo$^{1,2}$\quad Qing Guo$^{3}$ \quad Ka Chun Cheung$^{2, 4}$ \quad Simon See$^{2}$ \quad Renjie Wan$^{1}$\thanks{Corresponding author. This work was done at Renjie's Research Group at the Department of Computer Science of Hong Kong Baptist University.}\\
$^{1}$\small{Department of Computer Science, Hong Kong Baptist University}\\
$^{2}$\small{NVIDIA AI Technology Center, NVIDIA}\\
$^{3}$\small{IHPC and CFAR, Agency for Science, Technology and Research, Singapore}\\
$^{4}$\small{Department of Mathematics, Hong Kong Baptist University}\\
{\tt\small ziyuanluo@life.hkbu.edu.hk, guo_qing@cfar.a-star.edu.sg, \{chcheung, ssee\}@nvidia.com,}\\ {\tt\small renjiewan@hkbu.edu.hk}
}


\maketitle
\ificcvfinal\thispagestyle{empty}\fi

\begin{abstract}
Neural Radiance Fields (NeRF) have the potential to be a major representation of media. Since training a NeRF has never been an easy task, the protection of its model copyright should be a priority. In this paper, by analyzing the pros and cons of possible copyright protection solutions, we propose to protect the copyright of NeRF models by replacing the original color representation in NeRF with a watermarked color representation. Then, a distortion-resistant rendering scheme is designed to guarantee robust message extraction in 2D renderings of NeRF. Our proposed method can directly protect the copyright of NeRF models while maintaining high rendering quality and bit accuracy when compared among optional solutions. Project page: \href{https://luo-ziyuan.github.io/copyrnerf}{https://luo-ziyuan.github.io/copyrnerf}.


%
\end{abstract}

\section{Introduction}
\label{sec:intro}

\begin{figure}[t]
  \centering
  \includegraphics[width=\linewidth]{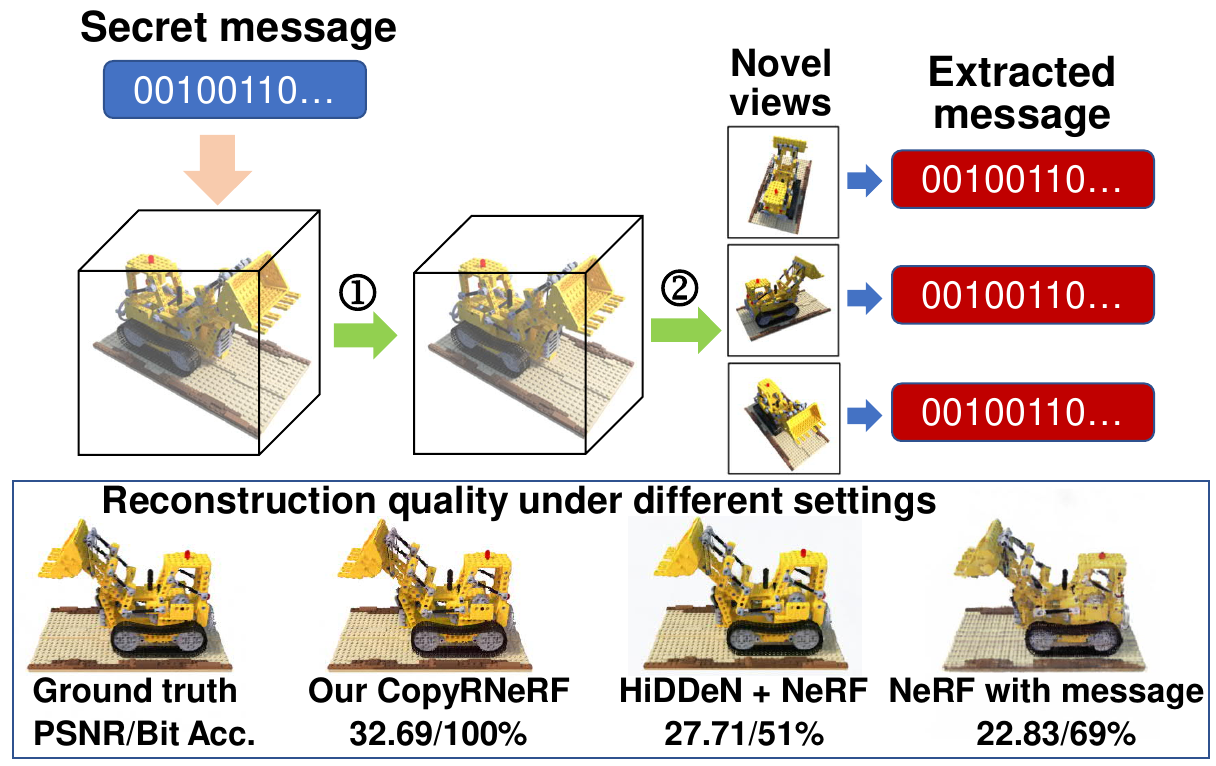}
  \caption{When NeRF models are stolen (\protect\circled{1}) by maclicious users, \CopyNeRF can help to claim model ownership by transmitting copyright messages embedded in models to rendering samples (\protect\circled{2}). We show some comparisons with HiDDeN~\cite{zhu2018hidden} + NeRF~\cite{nerf2020}, and NeRF~\cite{nerf2020} with messages. PSNR/Bit Accuracy is shown below each example.}
  \label{fig:short}
      \vspace{-15pt}
\end{figure}


Though Neural Radiance Fields (NeRF)~\cite{nerf2020} have the potential to be the mainstream for the representation of digital media, training a NeRF model has never been an easy task. If a NeRF model is stolen by malicious users, \textit{how can we identify its intellectual property?}

As with any digital asset (\eg, 3D model, video, or image), copyright can be secured by embedding copyright messages into asset, aka digital watermarking, and NeRF models are no exception. 
An intuitive solution is to directly watermark rendered samples using an off-the-shelf watermarking approach (\eg, HiDDeN~\cite{zhu2018hidden} and MBRS~\cite{jia2021mbrs}). However, this only protects the copyright of rendered samples, leaving the core model unprotected. If the core model has been stolen, malicious users may render new samples using different rendering strategies, leaving no room for external watermarking expected by model creators. Besides, without considering factors necessary for rendering during watermarking, directly watermarking rendered samples may leave easily detectable trace on areas with low geometry values.

The copyright messages are usually embedded into 3D structure (\eg, meshes) for explicit 3D models~\cite{Yoo_2022_CVPR}. Since such structures are all implicitly encoded into the weights of multilayer perceptron (MLP) for NeRF, its copyright protection should be conducted by watermarking model weights. As the information encoded by NeRF can only be accessed via 2D renderings of protected models, two common standards should be considered during the watermark extraction on rendered samples~\cite{ahmadi2020redmark, jing2021hinet, Yang_2021_ICCV, zhang2020udh}: 1) \textbf{invisibility}, which requires that no serious visual distortion are caused by embedded messages, and 2) \textbf{robustness}, which ensures robust message extraction even when various distortions are encountered.




One option is to create a NeRF model using watermarked images, while the popular invisible watermarks on 2D images cannot be effectively transmitted into NeRF models. As outlined in~\Fref{fig:short} (HiDDeN~\cite{zhu2018hidden} + NeRF~\cite{nerf2020}), though the rendered results are of high quality, the secret messages cannot be robustly extracted. We can also directly concatenate secret messages with input coordinates, which produces higher bit accuracy (NeRF with message in~\Fref{fig:short}). However, the lower PSNR values of rendered samples indicate that there is an obvious visual distortion, which violates the standard for invisibility. 

\begin{figure*}
  \centering
  \includegraphics[width=\linewidth]{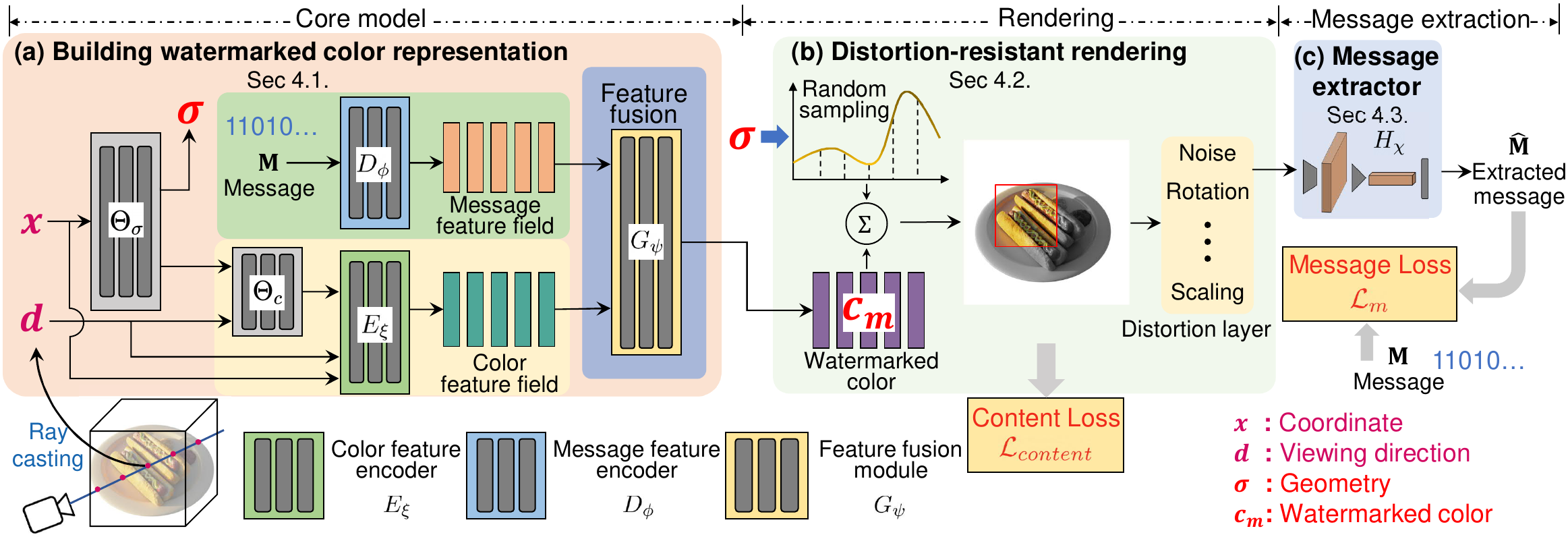}
  \caption{Illustration of our proposed method.  (a) A \textit{watermarked color representation} is obtained with the given secret message, which is able to produce watermarked color for rendering. (b) During training, a \textit{distortion-resistant rendering} is deployed to map the geometry ($\sigma$) and watermarked color representations to image patches with several distortions. (c) Finally, the secret message can be revealed by a CNN-based \textit{message extractor}. }
  

  \label{fig:framework}
    \vspace{-10pt}
\end{figure*}

Though invisibility is important for a watermarking system, the higher demand for robustness makes watermarking unique~\cite{zhu2018hidden}. Thus, in addition to invisibility, we focus on a more robust protection of NeRF models. As opposed to embedding messages into the entire models as in the above settings, we create a \textit{watermarked color representation} for rendering based on a subset of models, as displayed in~\Fref{fig:framework}. By keeping the base representation unchanged, this approach can produce rendering samples with invisible watermarks. By incorporating spatial information into the watermarked color representation, the embedded messages can remain consistent across different viewpoints rendered from NeRF models. We further strengthen the robustness of watermark extraction by using \textit{distortion-resistant rendering} during model optimization. A distortion layer is designed to ensure robust watermark extraction even when the rendered samples are severely distorted (\eg, blurring, noise, and rotation). A random sampling strategy is further considered to make the protected model robust to different sampling strategy during rendering. 

Distortion-resistant rendering is only needed during the optimization of core models. If the core model is stolen, even with different rendering schemes and sampling strategies, the copyright message can still be robustly extracted. Our contribution can be summarized as follows:

\begin{itemize}
    \item a method to produce copyright-embedded NeRF models.
    
    \item a watermarked color representation to ensure invisibility and high rendering quality.
    
    
    \item distortion-resistant rendering to ensure robustness across different rendering strategies or 2D distortions.
\end{itemize}

\section{Related work}

\paragraph{Neural radiance fields.} Various neural implicit scene representation schemes have been introduced recently \cite{niemeyer2020differentiable, yariv2020multiview, zhu2022neural}. The Scene Representation Networks (SNR) \cite{sitzmann2019scene} represent scenes as a multilayer perceptron (MLP) that maps world coordinates to local features, which can be trained from 2D images and their camera poses. DeepSDF \cite{park2019deepsdf} and DIST \cite{liu2020dist} use trained networks to represent a continuous signed distance function of a class of shapes. PIFu \cite{saito2019pifu} learned two pixel-aligned implicit functions to infer surface and texture of clothed humans respectively from a single input image. Occupancy Networks \cite{mescheder2019occupancy, peng2020convolutional} are proposed as an implicit representation of 3D geometry of 3D objects or scenes with 3D supervision. 
NeRF~\cite{nerf2020, zhu2023occlusion} in particular directly maps the 3D position and 2D viewing direction to color and geometry by a MLP and synthesize novel views via volume rendering. The improvements and applications of this implicit representation have been rapidly growing in recent years, including NeRF accelerating~\cite{fridovich2022plenoxels, neff2021donerf}, sparse reconstruction~\cite{yu2021pixelnerf, chen2021mvsnerf}, and generative models~\cite{schwarz2020graf, chan2021pi}. NeRF models are not easy to train and may use private data, so protecting their copyright becomes crucial.


\paragraph{Digital watermarking for 2D.} Early 2D watermarking approaches encode information in the least signiﬁcant bits of image pixels \cite{413536}. Some other methods instead encode information in the transform domains \cite{lai2010digital}. Deep-learning based methods for image watermarking have made substantial progress. HiDDeN~\cite{zhu2018hidden} was one of the first deep image watermarking methods that achieved superior performance compared to traditional watermarking approaches. RedMark~\cite{ahmadi2020redmark} introduced residual connections with a strength factor for
embedding binary images in the transform domain. Deep watermarking has since been generalized to video~\cite{weng2019high,zhang2019robust} as well. Modeling more complex and realistic image distortions also broadened the scope in terms of application~\cite{wengrowski2019light,tancik2020stegastamp}. However, those methods all cannot protect the copyright of 3D models.

\paragraph{Digital watermarking for 3D.} Traditional 3D watermarking approaches~\cite{ohbuchi2002frequency,praun1999robust,wu20153d} leveraged Fourier or wavelet analysis on triangular or polygonal meshes. Recently, Hou~\etal~\cite{hou2017blind} introduced a 3D watermarking method using the layering artifacts in 3D printed objects. Son~\etal~\cite{son2017perceptual} used mesh saliency as a perceptual metric to minimize vertex distortions. Hamidi~\etal~\cite{hamidi2019blind} further extended mesh saliency with wavelet transform to make 3D watermarking robust. Jing~\etal~\cite{liu2019novel} studied watermarking for point clouds through analyzing vertex curvatures. Recently, a deep-learning based approach \cite{Yoo_2022_CVPR} successfully embeds messages in 3D meshes and extracts them from 2D renderings. However, existing methods are for explicit 3D models, which cannot be used for NeRF models with implicit property.

\section{Preliminaries}
\label{sec:preliminary}
NeRF \cite{nerf2020} uses MLPs $\Theta_{\sigma}$ and $\Theta_{c}$ to map the 3D location $\mathbf{x} \in \mathbb{R}^3$ and viewing direction $\mathbf{d} \in \mathbb{R}^2$ to a color value $\mathbf{c} \in \mathbb{R}^3$ and a geometric value $\sigma \in \mathbb{R}^+$:
\begin{equation}
[\sigma, \mathbf{z}]=\Theta_{\sigma}\left(\gamma_{\mathbf{x}}(\mathbf{x})\right),
\label{eq:sigma}
\end{equation}
\begin{equation}
\mathbf{c}=\Theta_{c}\left(\mathbf{z}, \gamma_{\mathbf{d}}(\mathbf{d})\right),
\label{eq:nerffirst}
\end{equation}
where $\gamma_{\mathbf{x}}$ and $\gamma_{\mathbf{d}}$ are fixed encoding functions for location and viewing direction respectively. The intermediate variable $\mathbf{z}$ is a feature output by the first MLP $\Theta_{\sigma}$.


%
%

For rendering a 2D image from the radiance fields $\Theta_{\sigma}$ and $\Theta_{c}$, a numerical quadrature is used to approximate the volumetric projection integral. Formally, $N_p$ points are sampled along a camera ray $r$ with color and geometry values $\{(\mathbf{c}_r^i, \sigma_r^i)\}_{i=1}^N$. The RGB color value $\hat{\mathbf{C}}(r)$ is obtained using alpha composition
\begin{equation}
  \hat{\mathbf{C}}(r)=\sum_{i=1}^{N_p} T_r^i (1-\exp \left(-\sigma_r^i \delta_r^i\right)) \mathbf{c}_r^i,
  \label{eq:integral}
\end{equation}
where $T_r^i=\prod_{j=1}^{i-1}\left(\exp \left(-\sigma_r^i \delta_r^i\right)\right)$, and 
$\delta_r^i$ is the distance between adjacent sample points. The MLPs $\Theta_{\sigma}$ and $\Theta_{c}$ are optimized by minimizing a reconstruction loss between observations $\mathbf{C}$ and predictions $\hat{\mathbf{C}}$ as
\begin{equation}
  \mathcal{L}_{recon} = \frac{1}{N_r}\sum_{m =1}^{N_r}\|\hat{\mathbf{C}}(r_m)-\mathbf{C}(r_m)\|_2^2,
  \label{eq:recon}
\end{equation}
where $N_r$ is the number of sampling pixels. Given $\Theta_{\sigma}$ and $\Theta_{c}$, novel views can be synthesized by invoking volume rendering for each ray.

Considering the superior capability of NeRF in rendering novel views and representing various scenes, how can we protect its copyright when it is stolen by malicious users?



\section{Proposed method}

As outlined in~\Fref{fig:framework}, with a collection of 2D images $\{I_n\}_{n=1}^N$ and the binary message $\M \in \{0,1\}^{N_b}$ with length $N_b$, we address the issue raised in~\Sref{sec:preliminary} by building a watermarked color representation during optimization. In training, a distortion-resistant rendering is further applied to improve the robustness when 2D distortions or different rendering schemes are encountered. With the above design, the secret messages can be robustly extracted during testing even 
encountering sever distortions or different rendering strategies.

\subsection{Building watermarked color representation}
\label{sec:4.1}
The rendering in \Eref{eq:integral} relies on color and geometry produced by their corresponding representation in NeRF. To ensure the transmission of copyright messages to the rendered results, we propose embedding messages into their representation. We create a watermarked color representation on the basis of $\Theta_c$ defined in~\Eref{eq:nerffirst} to guarantee the message invisibility and consistency across viewpoints. The representation of geometry is also the potential for watermarking, but external information on geometry may undermine rendering quality~\cite{wang2022nerf, Huang22StylizedNeRF, chiang2022stylizing}. Therefore, the geometry does not become our first option, while experiments are also conducted to verify this setting.

We keep the geometry representation in~\Eref{eq:sigma} unchanged, and construct the watermarked color representation $\Theta_{m}$ to produce the message embedded color $\mathbf{c}_m$ as follows:
\begin{equation}
\mathbf{c}_m=\Theta_{m}\left(\mathbf{c},\gamma_{\mathbf{x}}(\mathbf{x}), \gamma_{\mathbf{d}}(\mathbf{d}), \mathbf{M}\right),
\end{equation}
where $\mathbf{M}$ denotes the message to be embedded and $\Theta_{m}$ contains several MLPs to ensure reliable message embedding. The input $\mathbf{c}$ is obtained by querying $\Theta_{c}$ using~\Eref{eq:nerffirst}. Several previous methods have pointed out the importance of building a 3D feature field when distributed features are needed to characterize 
composite information~\cite{xu20223d,chan2022efficient}. Thus, instead of directly fusing those information, we first construct their corresponding feature field and then combine them progressively.

\noindent\textbf{Color feature field.} In this stage, we aim at fusing the spatial information and color representation to ensure message consistency and robustness across viewpoints.  We adopt a color feature field by considering color, spatial positions, and viewing directions simultaneously as follows:
\begin{equation}
f_c=E_\xi(\mathbf{c},\gamma_{\mathbf{x}}(\mathbf{x}),\gamma_{\mathbf{d}}(\mathbf{d})).
\label{eq:colorrepresentation}
\end{equation}
Given a 3D coordinate $\mathbf{x}$ and a viewing direction $\mathbf{d}$, we first query the color representation $\Theta_{c}\left(\mathbf{z}, \gamma_{\mathbf{d}}(\mathbf{d})\right)$ to get $\mathbf{c}$, and then concatenate them with $\mathbf{x}$ and $\mathbf{d}$ to obtain spatial descriptor $\mathbf{v}$ as the input. Then the color feature encoder $E_{\xi}$ transforms $\mathbf{v}$ to the high-dimensional color feature field $f_c$ with dimension $N_c$. The Fourier feature encoding is applied to $\mathbf{x}$ and $\mathbf{d}$ before the feature extraction.

\noindent\textbf{Message feature field.} We further construct the message feature field. Specifically, we follow the classical setting in digital watermarking by transforming secret messages into higher dimensions~\cite{NIPS2017_838e8afb, baluja2019hiding}. It ensures more succinctly encoding of desired messages~\cite{NIPS2017_838e8afb}. As shown in~\Fref{fig:framework}, a message feature encoder is applied to map the messages to its corresponding higher dimensions as follows:
\begin{equation}
f_\M=D_{\phi}(\M).
\label{eq:messageembedding}
\end{equation}
In~\Eref{eq:messageembedding}, given message $\M$ of length $N_b$, the message feature encoder $D_{\phi}$ applies a MLP to the input message, resulting in a message feature field $f_\M$ of dimension $N_{m}$. 

Then, the watermarked color can be generated via a feature fusion module $G_{\psi}$ that integrates both color feature field and message feature field as follows: 
\begin{equation}
\mathbf{c}_m=G_{\psi}(f_c,f_\M,\mathbf{c}).
\label{eq:messagerepresentation}
\end{equation}
Specifically, $\mathbf{c}$ is also employed here to make the final results more stable. $\mathbf{c}_m$ is with the same dimension to $\mathbf{c}$, which ensures this representation can easily adapt to current rendering schemes.

\subsection{Distortion-resistant rendering}
\label{sec:4.2}
Directly employing the watermarked representation for volume rendering has already been able to guarantee invisibility and robustness across viewpoints. However, as discussed in~\Sref{sec:intro}, the message should be robustly extracted even when encountering diverse distortion to the rendered 2D images. Besides, for an implicit model relying on rendering to display its contents, the robustness should also be secured even when different rendering strategies are employed. Such requirement for robustness cannot be achieved by simply using  watermarked representation under the classical NeRF training framework. For example, the pixel-wise rendering strategy cannot effectively model the distortion (\eg, blurring and cropping) only meaningful in a wider scale. We, therefore, propose a distortion-resistant rendering by strengthening the robustness using a random sampling strategy and distortion layer.

Since most 2D distortions can only be obviously observed in a certain area, we consider the rendering process in a patch level~\cite{kwon2003watermarking, cotting2004robust}. A window with the random position is cropped from the input image with a certain height and width, then we uniformly sample the pixels from such window to form a smaller patch. The center of the patch is denoted by $\mathbf{u}=(u, v) \in \mathbb{R}^2$, and the size of patch is determined by $K\in \mathbb{R}^{+}$. We randomly draw the patch center $\mathbf{u}$ from a uniform distribution $\mathbf{u} \sim \mathcal{U}(\Omega)$ over the image domain $\Omega$. The patch $\mathcal{P}(\mathbf{u}, K)$ can be denoted by by a set of 2D image coordinates as
\begin{equation}
\mathcal{P}(\mathbf{u}, K)=\{(x+u, y+v) \mid x, y \in\{-\frac{K}{2}, \ldots, \frac{K}{2}-1\}\}.
\end{equation}
Such a patch-based scheme constitutes the backbone of our distortion-resistant rendering, due to its advantages in capturing information on a wider scale. Specifically, we employ a variable patch size to accommodate diverse distortions during rendering, which can ensure higher robustness in message extraction. This is because small patches increase the robustness against cropping attacks and large patches allow higher redundancy in the bit encoding, which leads to increased resilience against random noise~\cite{cotting2004robust}.

As the corresponding 3D rays are uniquely determined by $\mathcal{P}(\mathbf{u}, K)$, the camera pose and intrinsics, the image patch $\widetilde{\mathbf{P}}$ can be obtained after points sampling and rendering. Based on the sampling points in~\Sref{sec:preliminary}, we use a random sampling scheme to further improve the model's robustness, which is described as follows.

\noindent\textbf{Random sampling.} During volume rendering, NeRF~\cite{nerf2020} is required to sample 3D points along a ray to calculate the RGB value of a pixel color. However, the sampling strategy may vary as the renderer changes~\cite{neff2021donerf, lindell2021autoint}. To make our message extraction more robust even under different sampling strategies, we employ a random sampling strategy by adding a shifting value to the sampling points. Specifically, the original $N_p$ sampling points along ray $r$ is denoted by a sequence, which can be concluded as $\mathcal{X} = (x_r^1, x_r^2, \cdots, x_r^{N_p})$, where $x_r^{i},  i = 1, 2, \cdots, N_p$ denotes the sampling points during rendering. The randomized sample sequence $\mathcal{X}_{random}$ can be denoted by adding a shifting value as
\begin{equation}
\begin{aligned}
\mathcal{X}_{random} = (x_r^1 + z^1, x_r^2 + z^2, \cdots, x_r^{N_p} + z^{N_p}),\\
z^i \sim \mathcal{N}(0, \beta^2),\ i=1,2, \cdots, N_p,
\end{aligned}
\end{equation}
where $\mathcal{N}(0, \beta^2)$ is the Gaussian distribution with zero mean and standard deviation $\beta$.

By querying the watermarked color representation and geometry values at  $N_p$ points in $\mathcal{X}_{random}$, the rendering operator can be then applied to generate the watermarked color $\widetilde{\mathbf{C}}_m$ in rendered images:
\begin{equation}
  \widetilde{\mathbf{C}}_m(r)=\sum_{i=1}^{N_p} T_r^i (1-\exp \left(-\sigma_r^i \delta_r^i\right)) \mathbf{c}_m^i,
\end{equation}
where $T_r^i$ and $\delta_r^i$ are with the same definitions to their counterparts in~\Eref{eq:integral}.

All the colors obtained by  coordinates $\mathcal{P}$ can form a $K\times K$ image patch $\widetilde{\mathbf{P}}$. The content loss $\mathcal{L}_{content}$ of the 3D representation is calculated between watermarked patch $\widetilde{\mathbf{P}}$ and the $\hat{\mathbf{P}}$, where $\hat{\mathbf{P}}$ is rendered from the non-watermarked representation by the same coordinates $\mathcal{P}$.
In detail, the content loss $\mathcal{L}_{content}$ has two components namely pixel-wise MSE loss and perceptual loss:
\begin{equation}
  \mathcal{L}_{content} = \|\widetilde{\mathbf{P}}-\hat{\mathbf{P}}\|_2^2 + \lambda\|\Psi(\widetilde{\mathbf{P}}) - \Psi(\hat{\mathbf{P}})\|_2^2,
  \label{eq:content}
\end{equation}
where $\Psi(\cdot)$ denotes the feature representation obtained from a VGG-16 network, and $\lambda$ is a hyperparameter to balance the loss functions. 

\noindent\textbf{Distortion layer.}
To make our watermarking system robust to 2D distortions, a distortion layer is employed in our watermarking training pipeline after the patch $\widetilde{\mathbf{P}}$ is rendered. Several commonly used distortions are considered: 1) additive Gaussian noise with mean $\mu$ and standard deviation $\nu$; 2) random axis-angle rotation with parameters $\alpha$; and 3) random scaling with a parameter $s$; 4) Gaussian blur with kernel $k$. Since all these distortions are differentiable, we could train our network end-to-end. 

The distortion-resistant rendering is only applied during training. It is not a part of the core model. If the core model is stolen, even malicious users use different rendering strategy, the expected robustness can still be secured.

\subsection{Message extractor}
To retrieve message $\hat{\M}$ from the $K \times K$ rendered patch $\mathbf{P}$, a message extractor $H_{\chi}$ is proposed to be trained end-to-end:
\begin{equation}
H_{\chi}: \mathbb{R}^{K \times K}\rightarrow \mathbb{R}^{N_b},\ \mathbf{P} \mapsto \hat{\M},
\end{equation}
where $\chi$ is a trainable parameter. 
Specifically, we employ a sequence of 2D convolutional layers with the batch normalization and ReLU functions~\cite{ioffe2015batch}. An average pooling is then performed, following by a ﬁnal linear layer with a fixed output dimension $N_b$, which is the length of the message, to produce the continuous predicted message $\hat{\M}$. Because of the use of average pooling, the message extractor is compatible with any patch sizes, which means the network structure can remain unchanged when applying size-changing distortions such as random scaling.

The message loss $\mathcal{L}_m$ is then obtained by calculating the binary cross-entropy error between predicted message $\hat{\mathbf{M}}$ and the ground truth message $\mathbf{M}$:
\begin{equation}
  \mathcal{L}_m = {\rm mean}[-(\mathbf{M} \log \hat{\mathbf{M}}+(1-\mathbf{M}) \log (1-\hat{\mathbf{M}}))],
\end{equation}
where ${\rm mean}[\cdot]$ indicates the mean value over all bits.

To evaluate the bit accuracy during testing, the binary predicted message $\hat{\mathbf{M}}_b$ can be obtained by rounding:
\begin{equation}
\hat{\mathbf{M}}_b =  {\rm clamp}({\rm sign}(\hat{\mathbf{M}}), 0, 1),
\end{equation}
where $\rm clamp$ and $\rm sign$ are of the same definitions in~\cite{Yoo_2022_CVPR}. It should be noted that we use the continuous result $\hat{\mathbf{M}}$ in the training process, while the binary one $\hat{\mathbf{M}}_b$ is only adopted in testing process.

Therefore, the overall loss to train the copyright-protected neural radiance fields can be obtained as
\begin{equation}
\mathcal{L} = \gamma_1\mathcal{L}_{content} + \gamma_2\mathcal{L}_m,
\label{eq:loss}
\end{equation}
where $\gamma_1$ and $\gamma_2$ are hyperparameters to balance the loss functions.

\begin{figure*}
  \centering
  \includegraphics[width=\linewidth]{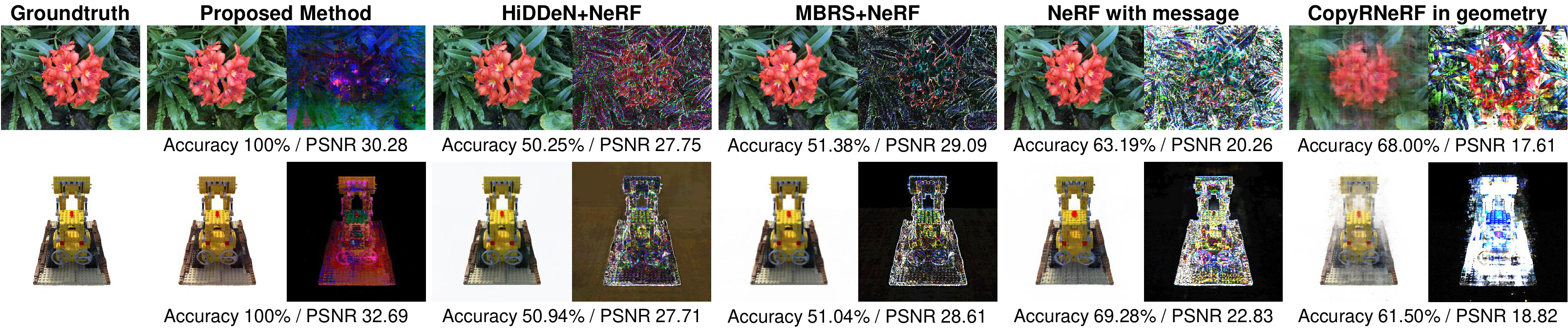}
  \caption{Visual quality comparisons of each baseline. We show the differences ($\times10$) between the synthesized results and the ground truth next to each method. Our proposed \CopyNeRF can achieve a well balance between the reconstruction quality and bit accuracy.} 
  \label{fig:visualquality}
  \vspace{-5pt}
\end{figure*}

\begin{table*}[!htb]
 \caption{Bit accuracies with different lengths compared with baselines. The results are averaged on all all examples.}
 \label{tab:comparison1}
 \centering
 \begin{tabular}{l|ccccc}
    \toprule
         & 4 bits & 8 bits & 16 bits & 32 bits & 48 bits \\
    \midrule
    \textbf{Proposed \CopyNeRF} &\textbf{100\%} & \textbf{100\%}  & \textbf{91.16\%} & \textbf{78.08\%}  & \textbf{60.06\%}   \\
    HiDDeN~\cite{zhang2020udh}+NeRF\cite{nerf2020} & 50.31\% & 50.25\% & 50.19\% & 50.11\% & 50.04\%      \\
    MBRS~\cite{jia2021mbrs}+NeRF~\cite{nerf2020} & 53.25\% & 51.38\% & 50.53\% & 49.80\% & 50.14\%      \\
    NeRF\cite{nerf2020} with message & 72.50\%  & 63.19\% &52.22\%  & 50.00\% & 51.04\%\\
    \CopyNeRF in geometry &   76.75\% & 68.00\% &60.16\%  & 54.86\% & 53.36\%\\
    \bottomrule
\end{tabular}
\vspace{-5pt}
\label{tab:bavsdl}
\end{table*}

\subsection{Implementation details}
We implement our method using PyTorch. An eight-layer MLP with $256$ channels and the following two MLP branches are used to predict the original colors $\mathbf{c}$ and opacities $\sigma$, respectively. We train a ``coarse'' network along with a ``fine'' network for importance sampling. we sample 32 points along each ray in the coarse model and 64 points in the fine model. Next, the patch size is set to $150\times 150$. The hyperparameters in~\Eref{eq:content} and~\Eref{eq:loss} are set as $\lambda_1=0.01$, $\gamma_1=1$, and $\gamma_2=5.00$. We use the Adam optimizer with defaults values $\beta_1 = 0.9$, $\beta_2 = 0.999$, $\epsilon = 10^{-8}$, and a learning rate $5\times10^{-4}$ that decays following the exponential scheduler during optimization. In our experiments, we set $N_m$ in \Eref{eq:messageembedding} as  $256$. We first optimize MLPs $\Theta_{\sigma}$ and $\Theta_{c}$ using loss function~\Eref{eq:recon} for $200$K and $100$K iterations for Blender dataset~\cite{nerf2020} and LLFF dataset~\cite{mildenhall2019local} separately, and then train the models $E_\xi$, $D_\phi$, and $H_\chi$ on 8 NVIDIA Tesla V100 GPUs. During training, we have considered messages with different bit lengths and forms. If a message has $4$ bits, we take into account all $2^{4}$ situations during training. The model creator can choose one message considered in our training as the desired message.




\section{Experiments}
\subsection{Experimental settings}
\noindent\textbf{Dataset.} To evaluate our methods, we train and test our model on Blender dataset~\cite{nerf2020} and LLFF dataset~\cite{mildenhall2019local}, which are common datasets used for NeRF. Blender dataset contains 8 detailed synthetic objects with 100 images taken from virtual cameras arranged on a hemisphere pointed inward. As in NeRF~\cite{nerf2020}, for each scene we input 100 views for training. LLFF dataset consists of 8 real-world scenes that contain mainly forward-facing images. Each scene contains 20 to 62 images. The data split for this dataset also follows NeRF~\cite{nerf2020}. For each scene, we select $20$ images from their testing dataset to evaluate the visual quality. For the evaluation of bit accuracy, we render $200$ views for each scene to test whether the message can be effectively extracted under different viewpoints. We report average values across all testing viewpoints in our experiments.

\noindent\textbf{Baselines.} To the best of our knowledge, there is no method specifically for protecting the copyright of NeRF models. We, therefore, compare with four strategies to guarantee a fair comparison: 1) \textbf{HiDDeN~\cite{zhu2018hidden}+NeRF\cite{nerf2020}}: processing images with classical 2D watermarking method HiDDeN~\cite{zhu2018hidden} before training the NeRF model; 2) \textbf{MBRS~\cite{jia2021mbrs}+NeRF~\cite{nerf2020}}: processing images with state-of-the-art 2D watermarking method MBRS~\cite{jia2021mbrs} before training the NeRF model; 3) \textbf{NeRF with message}: concatenating the message $\mathbf{M}$ with location $\mathbf{x}$ and viewing direction $\mathbf{d}$ as the input of NeRF; 4) \textbf{\CopyNeRF in geometry}: changing our \CopyNeRF by fusing messages with the geometry to evaluate whether geometry is a good option for message embedding. 



\noindent\textbf{Evaluation methodology.} We evaluate the performance of our proposed method against other methods by following the standard of digital watermarking about the invisibility, robustness, and capacity. For \textit{invisibility}, we evaluate the performance by using PSNR, SSIM, and LPIPS~\cite{zhang2018unreasonable} to compare the visual quality of the rendered results after message embedding. For \textit{robustness}, we will investigate whether the encoded messages can be extracted effectively by measuring the bit accuracy on different distortions. Besides normal situations, we consider the following distortions for message extraction: 1) Gaussian noise, 2) Rotation, 3) Scaling, and 4) Gaussian blur. For \textit{capacity}, following the setting in previous work for the watermarking of explicit 3D models~\cite{Yoo_2022_CVPR}, we investigate the invisibility and robustness under different message length as $N_b\in\{4, 8, 16, 32, 48\}$, which has been proven effective in protecting 3D models~\cite{Yoo_2022_CVPR}. Since we have included different viewpoints in our experiments for each scene, our evaluation can faithfully reflect whether the evaluated method can guarantee its robustness and consistency across viewpoints. 

\begin{figure*}
  \begin{minipage}{0.72\linewidth}
    \centering
    \tabcaption{Bit accuracies and reconstruction qualities compared with our baselines. $\uparrow$ ($\downarrow$) means higher (lower) is better. We show the results on $N_b = 16$  bits. The results are averaged on all all examples. The best performances are highlighted in \textbf{bold}.}
\label{tab:comparison2}
 \begin{tabular}{l|cccc}
    \toprule
         & Bit Acc$\uparrow$ & PSNR $\uparrow$& SSIM $\uparrow$& LPIPS $\downarrow$ \\
    \midrule
    \textbf{Proposed \CopyNeRF} &\textbf{91.16\%}& 26.29  & 0.910 & 0.038   \\
    HiDDeN~\cite{zhu2018hidden}+NeRF\cite{nerf2020} & 50.19\% & 26.53 & 0.917 & 0.035     \\
    MBRS~\cite{jia2021mbrs}+NeRF~\cite{nerf2020} & 50.53\% & \textbf{28.79} & \textbf{0.925} & \textbf{0.022}  \\
    NeRF with message & 52.22\% & 22.33 & 0.773 & 0.108\\
    \CopyNeRF in geometry & 60.16\% & 20.24 & 0.771  & 0.095 \\
    \bottomrule
\end{tabular}
\label{tab:bavq}
\end{minipage}
\begin{minipage}{0.28\linewidth}
    \centering
    \includegraphics[width = 0.76\linewidth]{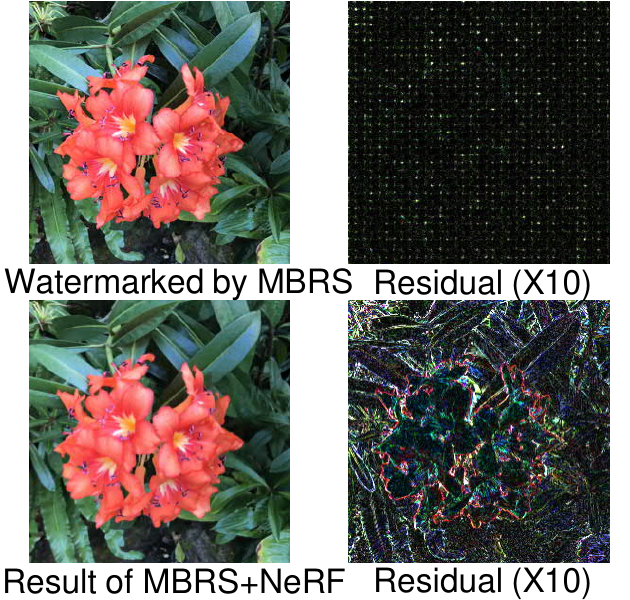}
    \vspace{-5pt}
    \figcaption{Analysis for failure of MBRS~\cite{jia2021mbrs}+NeRF.}
    \label{fig:compare}
  \end{minipage}
\end{figure*}

\begin{table*}[!htb]
 \caption{Bit accuracies with different distortion types compared with each baseline and our \CopyNeRF without distortion-resistant rendering (DRR).  We show the results on $N_b = 16$ bits. The results are averaged on all all examples.}
 \label{tab:comparison3}
 \centering
 \begin{tabular}{l|ccccc}
    \toprule
         & No Distortion & Gaussian noise  & Rotation  & Scaling & Gaussian blur\\
         & & ($\nu$=0.1) & ($\pm\pi/6$) & ($\le25\%$) & ($deviation = 0.1$) \\
    \midrule
    \textbf{Proposed \CopyNeRF} &91.16\% & \textbf{90.44\%}  & \textbf{88.13\%} & \textbf{89.33\%} &  \textbf{90.06\%}   \\
    HiDDeN~\cite{zhu2018hidden}+NeRF\cite{nerf2020} & 50.19\% & 49.84\% & 50.12\% & 50.09\% &  50.16\%    \\
    MBRS~\cite{jia2021mbrs}+NeRF~\cite{nerf2020} &  50.53\% & 51.00\%  & 51.03\% & 50.12\% & 50.41\% \\
    NeRF with message & 52.22\%  & 50.53\%       & 50.22\%  & 50.19\% & 51.34\%\\
    \CopyNeRF in geometry &   60.16\% & 58.00\% &56.94\%  & 60.09\% & 59.38\%\\
    \CopyNeRF W/o DRR & \textbf{91.25\%}  & 89.12\%  & 75.81\%  & 87.44\% & 87.06\%\\
    \bottomrule
\end{tabular}
\vspace{-5pt}
 \label{tab:distortion}
 \end{table*}

\subsection{Experimental results}
\noindent\textbf{Qualitative results.}
\label{sec:qualitativeresults}
We first compare the reconstruction quality visually against all baselines and the results are shown in~\Fref{fig:visualquality}. Actually, all methods except NeRF with message and \CopyNeRF in geometry can achieve high reconstruction quality. For HiDDeN~\cite{zhu2018hidden} + NeRF~\cite{nerf2020} and MBRS~\cite{jia2021mbrs}+NeRF~\cite{nerf2020}, although they are efficient approaches in 2D watermarking, their bit accuracy values are all low for rendered images, which proves that the message are not effectively embedded after NeRF model training. From the results shown in \Fref{fig:compare}, the view synthesis of NeRF changes the information embedded by 2D watermarking methods, leading to their failures. For NeRF with message, as assumed in our previous discussions, directly employing secret messages as an input change the appearance of the output, which leads to their lower PSNR values. Besides, its lower bit accuracy also proves that this is not an effective embedding scheme. For \CopyNeRF in geometry, it achieves the worst visual quality among all methods. The rendered results look blurred, which confirms our assumption that the geometry is not a good option for message embedding.

\noindent\textbf{Bit Accuracy vs. Message Length.}
We launch $5$ experiments for each message length and show
the relationship between bit accuracy and the length of message in~\Tref{tab:bavsdl}. We could clearly see that the bit accuracy drops when the number of bits increases. However, our \CopyNeRF achieves the best bit accuracy across all settings, which proves that the messages can be effectively embedded and robustly extracted. \CopyNeRF in geometry achieves the second best results among all setting, which shows that embedding message in geometry should also be a potential option for watermarking. However, the higher performance of our proposed \CopyNeRF shows that color representation is a better choice. 


\begin{figure*}
  \centering
  \includegraphics[width=\linewidth]{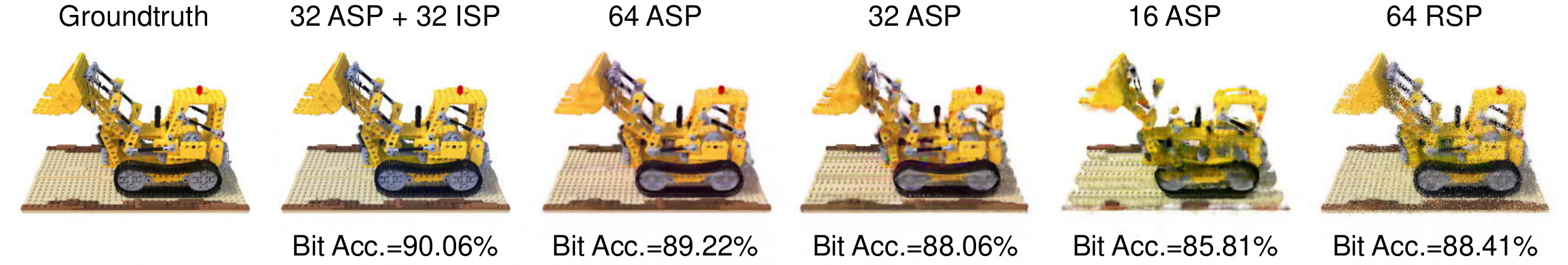}
  \caption{Comparisons for different rendering degradadtion in the inference phase. The message length is set to 16. We use average sampling points (ASP), importance sampling points (ISP), and random sampling points (RSP) in different rendering strategies. ``$32$ ASP + $32$ ISP'' is a strategy employed in the training process, and message extraction also shows the highest bit accuracy. When sampling strategies are changed to other ones during inference, the message extraction still shows similar performance, which verifies the effectiveness of our distortion-resistant rendering.}
  \label{fig:diff_render}
  \vspace{-12pt}
\end{figure*}

\noindent\textbf{Bit Accuracy vs. Reconstruction Quality.}
 We conduct more experiments to evaluate the relationship between bit accuracy and reconstruction quality. The results are shown in~\Tref{tab:bavq}\footnote{Results for other lengths of raw bits can be found in the supplementary materials.}. Our proposed \CopyNeRF achieves a good balance between bit accuracy and error metric values. Though the visual quality values are not the highest, the bit accuracy is the best among all settings. Though HiDDeN~\cite{zhu2018hidden} + NeRF~\cite{nerf2020} and MBRS~\cite{jia2021mbrs}+NeRF~\cite{nerf2020} can produce better visual quality values, its lower bit accuracy indicates that the secret messages are not effectively embedded and robustly extracted. NeRF with message also achieves degraded performance on bit accuracy, and its visual quality values are also low. It indicates that the embedded messages undermine the quality of reconstruction. Specifically, the lower visual quality values of \CopyNeRF in geometry indicates that hiding messages in color may lead to better reconstruction quality than hiding messages in geometry.




\noindent\textbf{Model robustness on 2D distortions.}
We evaluate the robustness of our method by applying several traditional 2D distortions. Specifically, as shown in~\Tref{tab:distortion}, we consider several types of 2D distortions including noise, rotation, scaling, and cropping. We could see that our method is quite robust to different 2D distortions. Specifically, \CopyNeRF w/o DRR achieves similar performance to the complete \CopyNeRF when no distortion is encountered. However, when it comes to different distortions, its lower bit accuracies demonstrate the effectiveness of our distortion-resistant rendering during training.

\noindent\textbf{Analysis for feature field.}
\label{sec:5.3}
In the section, we further evaluate the effectiveness of color feature field and message feature field. We first remove the module for building color feature field and directly combine the color representation with the message features. In this case, the model performs poorly in preserving the visual quality of the rendered results. We further remove the module for building message feature field and combine the message directly with the color feature field. The results in~\Tref{tab:ablationstudy} indicate that this may result in lower bit accuracy, which proves that messages are not embedded effectively. 

\begin{table}[tb]
 \caption{Comparisons for our full model, our model without Message Feature Field (MFF) and our model without Color Feature Field (CFF). The last row shows that our method achieves consistent performance even when different rendering scheme (DRS) is applied during \underline{testing}. }
 \label{tab:comparison4}
 \centering
 \begin{tabular}{l|cccc}
    \toprule
         &Bit Acc$\uparrow$& PSNR $\uparrow$& SSIM $\uparrow$& LPIPS $\downarrow$ \\
    \midrule
    \textbf{Ours} & \textbf{100\%} & \textbf{32.68} & \textbf{0.948} & \textbf{0.048}  \\
    W/o MFF& 82.69\% & 20.46 & 0.552 & 0.285     \\
    W/o CFF& 80.69\% & 21.06 & 0.612 & 0.187     \\
    DRS  & \textbf{100\%}  & 32.17 & 0.947 & 0.052   \\
    \bottomrule
\end{tabular}
\label{tab:ablationstudy}
      \vspace{-12pt}
 \end{table}
\noindent\textbf{Model robustness on rendering.}
Though we apply a normal volume rendering strategy for inference, the messages can also be effectively extracted using a distortion rendering utilized in training phase. As shown in the last row of~\Tref{tab:ablationstudy}, the quantitative values with the distortion rendering are still similar to original results in the first row of~\Tref{tab:ablationstudy}, which further confirms the robustness of our proposed method.

The results for different sampling schemes are presented in~\Fref{fig:diff_render}. Our distortion-resistant rendering employs $32$ average sampling points and $32$ importance sampling points during training. When different sampling strategies are applied in the inference phase, our method can also achieve high bit accuracy, which can validate the robustness of our method referring to different sampling strategies.

\begin{figure}
  \centering
  \includegraphics[width=\linewidth]{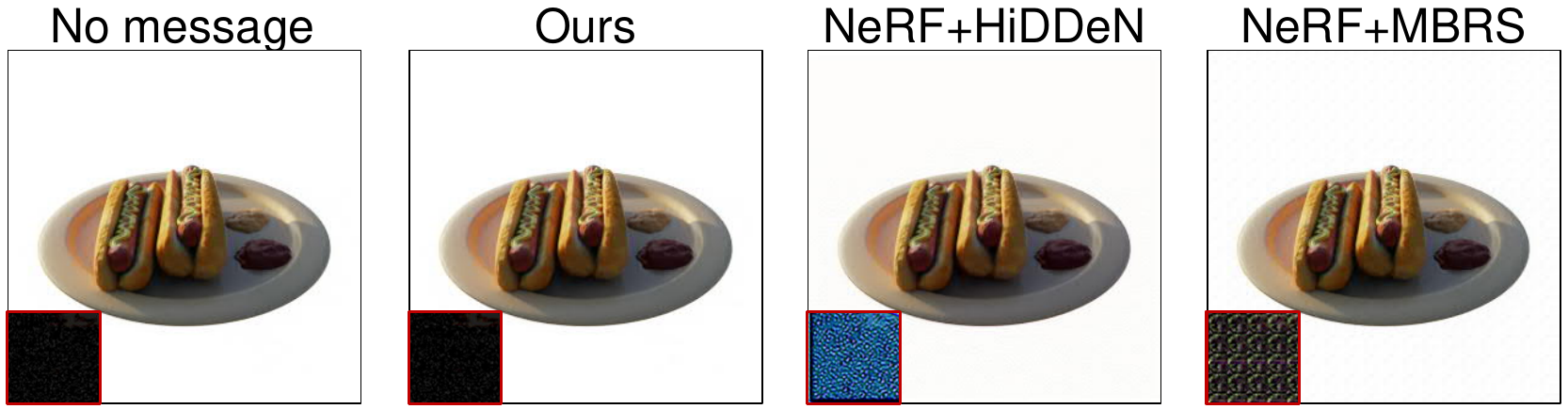}
  \caption{Comparisons for watermarking after rendering. The patch in the lower left corner shows the augmentation result by simply multiplying a factor 30. We use image inversion for better visualization}
  \label{fig:hotdog}
  \vspace{-15pt}
\end{figure}

\noindent\textbf{Comparison with NeRF+HiDDeN/MBRS~\cite{zhu2018hidden, jia2021mbrs}.} We also conduct an experiment to compare our method with approaches by directly applying 2D watermarking method on rendered images, namely NeRF+HiDDeN~\cite{zhu2018hidden} and NeRF+MBRS~\cite{jia2021mbrs}. Although these methods can reach a high bit accuracy as reported in their papers, 
as shown in \Fref{fig:hotdog}, these methods can easily leave detectable traces especially in areas with lower geometry values, as they lack the consideration for 3D information during watermarking. Besides, they only consider the media in 2D domain and cannot protect the NeRF model weights.
\section{Conclusions}

In this paper, we propose a framework to create a copyright-embedded 3D implicit representation by embedding messages into model weights. In order to guarantee the invisibility of embedded information, we keep the geometry unchanged and construct a watermarked color representation to produce the message embedded color. The embedded message can be extracted by a CNN-based extractor from rendered images from any viewpoints, while keeping high reconstruction quality. Additionally, we introduce a distortion-resistant rendering scheme to enhance the robustness of our model under different types of distortion, including classical 2D degradation and different rendering strategies. The proposed method achieves a promising balance between bit accuracy and high visual quality in experimental evaluations.

\noindent\textbf{Limitations.} Though our method has shown promising performance in claiming the ownership of Neural Radiance Fields,  
training a NeRF model is time-consuming. We will consider how to speed up the training process in our future work. Besides, though we have considered several designs to strengthen the system robustness, this standard may still be undermined when malicious users directly attack model weights, \ie, the model weights are corrupted. We conduct a simple experiment by directly adding Gaussian noise (std = 0.01) to the model parameters, and the accuracy slightly decreases to $93.97\%$ ($N_b = 8$). As this may also affect rendering quality, such model weights corruption may not be a priority for malicious users who intend to display the content. We will still actively consider how to handle such attacks in our future work. 
 
\noindent\textbf{Acknowledgement.}  Renjie Wan is supported by the Blue Sky Research Fund of HKBU under Grant No. BSRF/21-22/16 and Guangdong Basic and Applied Basic Research Foundation under Grant No. 2022A1515110692. Qing Guo is supported by the A*STAR Centre for Frontier AI Research and the National Research Foundation, Singapore, and DSO National Laboratories under the AI Singapore Programme (AISG Award No: AISG2-GC-2023-008).

\linespread{1.0}
\normalem
{\small
\bibliographystyle{ieee_fullname}
\bibliography{egbib}
}

\end{document}